\documentclass[fleqn,10pt]{wlscirep}\twocolumn
\usepackage[utf8]{inputenc}
\usepackage[T1]{fontenc}
\usepackage{amssymb}
\usepackage{booktabs}
\usepackage{bbding}

\usepackage{threeparttable}

\title{GP-VLS: A general-purpose vision language model for surgery}

\author[1,2,†*]{Samuel Schmidgall}
\author[1,†]{Joseph Cho}
\author[1]{Cyril Zakka}
\author[1]{William Hiesinger}
\affil[1]{Department of Cardiothoracic Surgery, Stanford University}
\affil[2]{Department of Electrical and Computer Engineering, Johns Hopkins University}
 
\affil[$\dagger$]{ Equal Contribution}
\affil[*]{sschmi46@jhu.edu}


\begin{abstract}        

Surgery requires comprehensive medical knowledge, visual assessment skills, and procedural expertise. While recent surgical AI models have focused on solving task-specific problems, there is a need for general-purpose systems that can understand surgical scenes and interact through natural language. This paper introduces GP-VLS, a general-purpose vision language model for surgery that integrates medical and surgical knowledge with visual scene understanding. For comprehensively evaluating general-purpose surgical models, we propose SurgiQual, which evaluates across medical and surgical knowledge benchmarks as well as surgical vision-language questions. To train GP-VLS, we develop six new datasets spanning medical knowledge, surgical textbooks, and vision-language pairs for tasks like phase recognition and tool identification. We show that GP-VLS significantly outperforms existing open- and closed-source models on surgical vision-language tasks, with 8-21\% improvements in accuracy across SurgiQual benchmarks. GP-VLS also demonstrates strong performance on medical and surgical knowledge tests compared to open-source alternatives. Overall, GP-VLS provides an open-source foundation for developing AI assistants to support surgeons across a wide range of tasks and scenarios. The code and data for this work is publicly available at \href{https://gpvls-surgery-vlm.github.io/}{gpvls-surgery-vlm.github.io}.
 
\end{abstract} 
\begin{document}

\flushbottom
\maketitle

\thispagestyle{empty}

\section{Introduction}

\begin{figure*}
    \centering    \includegraphics[width=1.0\linewidth]{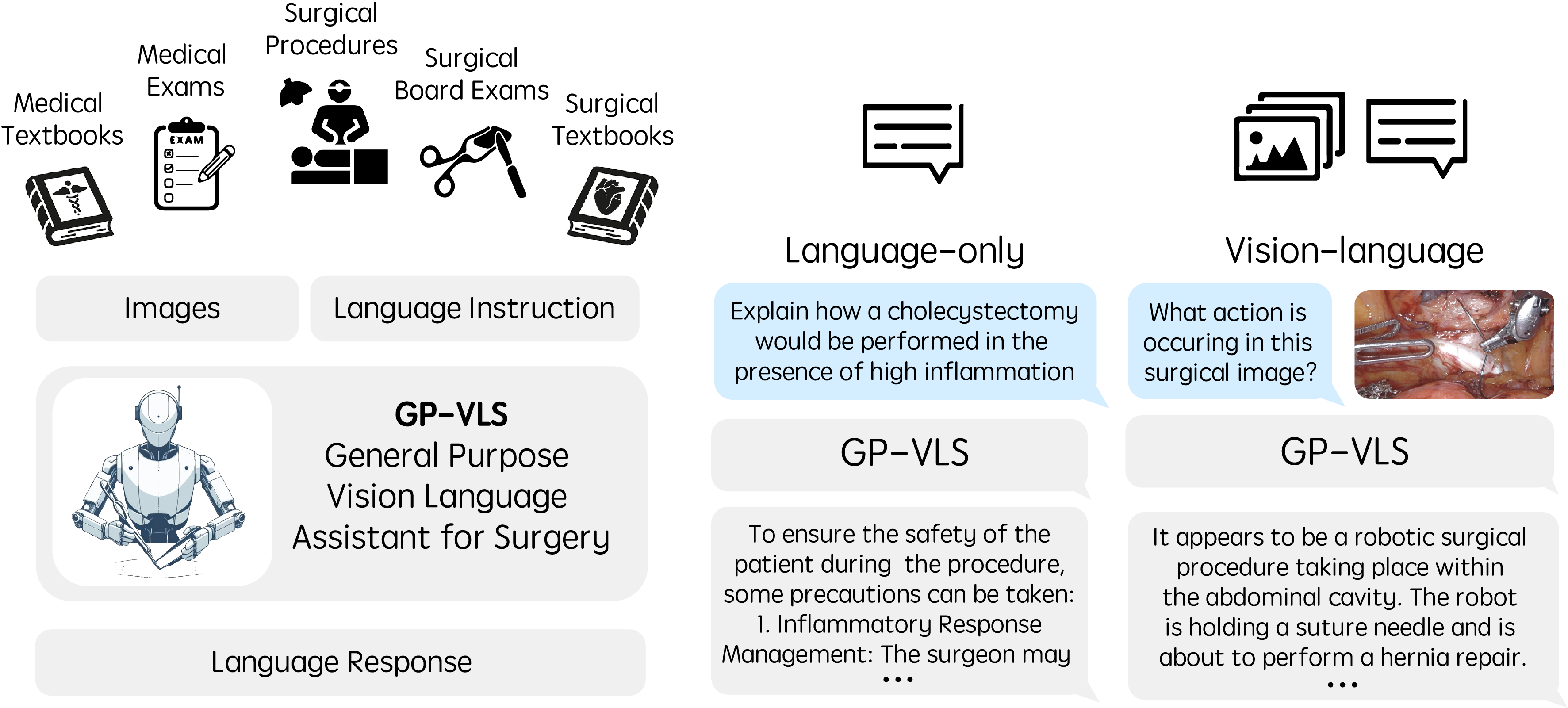}
    \caption{Visual depiction of the General Purpose Vision Language Assistant for Surgery (GP-VLS) and the content used to train it. GP-VLS is trained on and is able to perform language-only and vision-language problems.}
    \label{VLM}
\end{figure*}
           
Surgery demands a complex interplay of skills, combining comprehensive medical knowledge, visual assessment capabilities, and procedural expertise. 
The emerging field of surgical AI offers exciting possibilities to augment surgeons' capabilities and potentially transform surgical practice\cite{hashimoto2018artificial,pakkasjarvi2023artificial,varghese2024artificial}. One of the primary goals in this field is to develop general-purpose systems capable of understanding surgical scenes and interacting with clinicians through natural language\cite{schmidgall2024general,varghese2024artificial}. Such systems could revolutionize various aspects of surgery, from preoperative planning to intraoperative guidance and postoperative care.
     
The potential impact of these AI assistants extends beyond simple task automation. By integrating broad medical knowledge with real-time visual understanding of surgical scenes, these systems could serve as surgical collaborators that offer context-aware insights and decision support during critical moments in the operating room\cite{varghese2024artificial}. Such a model could also provide a foundation for future robotic surgery applications where AI systems can not only execute tasks but also explain their reasoning in human-understandable terms\cite{schmidgall2024general, schmidgall2024robots}. For surgeons, this explainability is crucial, as it allows for meaningful oversight and collaboration with AI systems, ensuring that the technology augments rather than replaces human expertise. 

Recently, vision language models (VLMs) have made significant progress in open-world visual understanding\cite{liu2024visual, liu2024improved,liu2024llavanext}, image captioning\cite{zhou2020unified,hu2022scaling,yang2023vid2seq}, and visual question answering\cite{shao2023prompting, guo2023images, salaberria2023image}. 
The idea of VLMs is to integrate data from multiple sensory modalities, including text, image, and sometimes audio, to create a more comprehensive understanding of input data\cite{zhang2024vision}. This integration allows VLMs to perform complex tasks across different domains more effectively than unimodal systems, which are limited to a single type of input.
Additionally, VLMs typically produce text as output, which enables them to generate descriptions, explanations, or other forms of textual content analysis that humans can easily understand.

In addition to general-purpose VLMs, many VLMs have been proposed for various medical specialties, such as pathology\cite{lu2023foundational, lu2024visual,huang2023visual}, radiology\cite{pellegrini2023radialog,chambon2022roentgen}, and general medicine\cite{alayrac2022flamingo}. Several VLMs have also been proposed for surgery, showing high performance on surgical subtasks such as tool recognition and phase labelling\cite{seenivasan2023surgicalgpt, wang2024surgical, bai2023surgical}. The typical training workflow for surgical VLMs is either converting existing classification datasets into language datasets and training a new model from scratch\cite{wang2024surgical}, or attaching a classification head and not producing language all together\cite{seenivasan2023surgicalgpt, bai2023surgical}. However, because the datasets of the models outputting language dataset are derived from classification labels their complexity is relatively low and is often structured in the form of yes or no responses. As a result, these models perform well on the proposed benchmarks but struggle with general language capabilities. Additionally, these models do not have an underlying technical knowledge of medicine or surgery--when provided with medical knowledge benchmarks these models are not able to correctly answer \textit{any} questions.

In this study, we introduce a general-purpose surgical VLM which is able to understand fundamental concepts in medicine and surgery as well as surgical vision-language problems. We propose a comprehensive quality metric for surgical VLMs which assesses the models' proficiency in medical knowledge, surgical knowledge, and surgical procedures. The code and data for this work is publicly available at \href{https://gpvls-surgery-vlm.github.io/}{gpvls-surgery-vlm.github.io}. The main contributions in this work are as follows:

\begin{enumerate}
    \item An open-source general-purpose vision language model for surgery (GP-VLS) which is able to understand fundamental concepts in medicine and surgery as well as surgical vision-language problems.
    \item A comprehensive quality metric for surgical vision-language problems in order to assess both knowledge of medicine and surgery as well as an understanding of surgical scenes.
    \item Six new surgery training datasets across a wide variety of tasks. Five of the introduced datasets focus on vision-language, including understanding surgical scenes such as surgical triplets or tool locations. We also introduce a novel training dataset derived from surgical textbooks with an emphasis on general and cardiothoracic surgery.

\end{enumerate}

\section{Related works}

\subsection{Visual Instruction Tuning}

\paragraph{Instruction following models} In computer vision, designing models that follow instructions typically result from two design pathways: \textbf{(i)} The first pathways produces models that are trained to perform a particular operation end-to-end on labelled data. These types of models are typically trained to solve problems that are specific to the task. Examples of this include models which perform classification\cite{redmon2016you, redmon2017yolo9000, bochkovskiy2020yolov4} or segmentation\cite{kirillov2023segment,ke2024segment,mazurowski2023segment}, with the final model is only able to perform those tasks. \textbf{(ii)} The second pathway aims to produce models that are generally capable and can solve a variety of problems. This pathway typically coordinates various models through a language interface, such as VLMs\cite{liu2024visual, liu2024improved,liu2024llavanext}.

\paragraph{Instruction tuning} In order to enable language models and VLMs to follow instructions in natural language, a technique called \textit{instruction tuning} is often used, such as LLaVA\cite{liu2024visual}, InstructGPT\cite{ouyang2022training}, and FLAN-PaLM\cite{chung2024scaling}. Instruction tuning is a technique to improve the performance of a pre-trained model by fine-tuning on a curated dataset focused on accurately executing specific instructions\cite{zhang2023instruction}. Instruction tuning improves traditional fine-tuning by aligning the model's outputs with the focused intentions of the instructions. For example, a model could be instruction-tuned with a dataset pairing the command "Summarize this article" with concise, well-crafted summaries to teach it to generate accurate article summaries. The goal is to refine the model’s ability to generate relevant, precise responses. While this approach is technically quite simple, it has been shown that instruct-tuning is effective for improving zero- and few-shot generalization capabilities\cite{liu2024llavanext}.

\begin{table*}[ht]
\centering
\begin{tabular}{@{}l|cc|cc|cccc@{}}
\toprule
\textbf{Model} & \textbf{Medical} & \textbf{Surgical} & \textbf{\# Tasks} & \textbf{\# Datasets} & \textbf{Language} & \textbf{Open weights}  & \textbf{Open code} & \textbf{Inference} \\ 
\midrule
Surgical-LVLM & \color{red}{\XSolidBrush} & $\Large\color{green}{\checkmark}$ & 1 & 3 & $\Large\color{green}{\checkmark}$ & \color{red}{\XSolidBrush} & $\Large\color{green}{\checkmark}$ & \color{red}{\XSolidBrush}\\
SurgicalGPT & \color{red}{\XSolidBrush} & $\Large\color{green}{\checkmark}$ & 3 & 3 & \color{red}{\XSolidBrush} & \color{red}{\XSolidBrush} & $\Large\color{green}{\checkmark}$ & \color{red}{\XSolidBrush}\\
Surgical-VQA & \color{red}{\XSolidBrush} & $\Large\color{green}{\checkmark}$ & 3 & 3 & \color{red}{\XSolidBrush} & $\Large\color{green}{\checkmark}$ & $\Large\color{green}{\checkmark}$ & $\Large\color{green}{\checkmark}$\\
Surgical-VQLA & \color{red}{\XSolidBrush} & $\Large\color{green}{\checkmark}$ & -- & -- & \color{red}{\XSolidBrush} & \color{red}{\XSolidBrush} & $\Large\color{green}{\checkmark}$ & \color{red}{\XSolidBrush}\\
Med-Gemini & $\Large\color{green}{\checkmark}$ & \large\color{red}{\textbf{--}} & \large\color{red}{\textbf{--}} & \large\color{red}{\textbf{--}} & $\Large\color{green}{\checkmark}$ & \color{red}{\XSolidBrush} & \color{red}{\XSolidBrush} & \color{red}{\XSolidBrush}\\
GPT-4 & $\Large\color{green}{\checkmark}$ & \large\color{red}{\textbf{--}} & \large\color{red}{\textbf{--}} & \large\color{red}{\textbf{--}} & $\Large\color{green}{\checkmark}$ & \color{red}{\XSolidBrush} & \color{red}{\XSolidBrush} & $\Large\color{green}{\checkmark}$\\
GPT-4 Omni & $\Large\color{green}{\checkmark}$ & \large\color{red}{\textbf{--}} & \large\color{red}{\textbf{--}} & \large\color{red}{\textbf{--}} & $\Large\color{green}{\checkmark}$ &  \color{red}{\XSolidBrush} & \color{red}{\XSolidBrush} & $\Large\color{green}{\checkmark}$\\
\toprule
\textbf{GP-VLS (ours)} & $\Large\color{green}{\checkmark}$ & $\Large\color{green}{\checkmark}$ & \textbf{7} & \textbf{10} & $\Large\color{green} {\checkmark}$ & $\Large\color{green}{\checkmark}$ & $\Large\color{green}{\checkmark}$ & $\Large\color{green}{\checkmark}$\\
\bottomrule
\end{tabular}
\caption{Comparison of various vision-language models trained on either surgical or medical data.}
\label{datasets}
\end{table*}

\subsection{Vision Language Models}

VLMs are language models designed to process both visual and textual information. These models combine LLMs with vision processing to enable interactions between image and text data using a language backbone\cite{zhang2024vision}. Typically, VLMs consist of a vision encoder $f_v(I)$ for processing images, a text encoder $f_t(T)$ for handling language inputs, and a multimodal fusion component $f_m(v,t)$ to integrate information from both sources. The vision encoder often employs convolutional neural networks or transformer-based architectures to extract features, while the text encoder typically uses transformer models. 

The fusion of visual and textual features can be achieved through simple concatenation or more complex attention mechanisms:
$f_m(v,t) = \text{Attention}(v,t) = \text{softmax}(vt^T)t$
where $v$ and $t$ are the visual and textual features respectively. One of the most common pre-training strategy is contrastive learning which utilizes a loss function as follows:
$L = -\log \frac{\exp(s(v, t))}{\sum_{j} \exp(s(v, t_j))}$
where $s(v,t)$ is the similarity between matched image-text pairs, and the sum is over all text samples in a batch. See ref.\cite{liu2024visual} for a more in-depth discussion of this.

\paragraph{VLMs in medicine} There have been several VLMs proposed for use in medicine. 
\cite{moor2023med}
Med-Gemini is a general VLM built on Gemini-1.5 which focuses on solving a broad range of applications across various medical domains\cite{saab2024capabilities}.
Almanac is a retrieval-augmented language model framework designed to improve factual correctness in clinical decision-making by incorporating external knowledge sources\cite{zakka2024almanac}.
While not technically a medical-specialized VLM, GPT-4 has performed as one of the most accurate VLMs on medical benchmarks\cite{nori2023can, schmidgall2024addressing,ziaei2023language, nori2023capabilities}. GPT-4 Omni has also shown high performance on QA and multimodal medical problems\cite{schmidgall2024agentclinic}. Performance with general models, such as GPT-4, has recently been shown to dramatically improve using prompting techniques, such as ensemble and chain-of-thought\cite{nori2023can}.

\paragraph{VLMs in surgery}
Several VLMs have been proposed specifically for surgery.
SurgicalGPT\cite{seenivasan2023surgicalgpt} is a VLM that integrates vision and language processing to respond to questions based on surgical scenes. It employs a hybrid architecture combining GPT models with visual feature extraction to perform classification on three tasks: triplet-pair, phase, and step classification. Surgical-VQA\cite{seenivasan2022surgical} is designed for VQA within surgical videos, which uses transformer-based architectures to understand and answer questions about surgical tools, procedures, and interactions. This model is designed to provide support in surgical training and real-time surgery by interpreting complex visual scenes and textual queries. 

Surgical-VQLA\cite{bai2023surgical} performs classification for surgical action and also localizes the relevant areas in the image, using a Gated Vision-Language Embedding technique. This model eliminates the need for a separate object detection step, making it faster and more suitable for real-time applications in surgical training and understanding. 
Surgical-LVLM\cite{wang2024surgical} focuses on the integration of detailed visual analysis with language-based interaction. To the best of our knowledge, this is the only surgical language model which outputs text instead of a classification label outside of our work. VLMs have also been used for facilitating surgical sub-task implementations\cite{moghani2024sufia}, outputting sub-task commands for a surgical robot to follow.

\paragraph{Comparing surgical \& medical VLMs} In Table \ref{datasets} we demonstrate an outline of various surgical and medical models based on various supported parameters. Of the four currently existing surgical VLMs, only one has language as output while the other three use classification labels on specific tasks as output. Additionally, none of the closed-source models were trained on medical data (only on three surgical datasets each) and only one work open-sourced their model weights to allow for comparing performance. For the three "medical" models the three closed source models (Med-Gemini\cite{saab2024capabilities}, GPT-4\cite{achiam2023gpt}, and GPT-4 Omni) do not share weights or code, and Med-Gemini does not even allow access to inference for all researchers. It is also not clear what data was used to train these models and whether this includes surgery.

\begin{table*}[ht]
    \centering
    \begin{threeparttable}
    \begin{tabular}{lcccccc}
        \toprule
        Model & MedQA & MedMCQA-Surgery & Phase Recgn & Triplet Recgn & Tool Recgn & Action Recgn \\
        \midrule
        prism-clip+7b & 32.3 & 33.1 & 13.2 & 1.1 & 39.5 & 32.7 \\
               prism-clip+13b & 34.6 & 37.9 & 9.4 & 3.9 & 36.7 & 25.2  \\
                dinosiglip+7b & 34.2 & 36.0 & 14.0 & 16.6 & 36.1 & 32.5  \\
               dinosiglip+13b & 34.3 & 36.3 & 11.2 & 2.3 & 34.4 & 26.5  \\
               \hline
        GPT-4 Omni & 80.5 & \textbf{74.3} & 31.6 & 14.3 & 79.9 & 31.0 \\
        GPT-4 & \textbf{86.1} & 62.6 & 30.6 & 7.4 & 79.1 & 30.5  \\
        \midrule
        GP-VLS (ours) & 46.1 & 52.8 & \textbf{39.8} & \textbf{37.3} & \textbf{94.4} & \textbf{49.6} \\
        
        \bottomrule
    \end{tabular}
    \caption{Performance of vision-language models on the SurgiQual benchmark.}
    \label{results}
    \end{threeparttable}
\end{table*}

\section{Results}

\subsection{Training Datasets}

In order to train a generalist surgical VLM, we introduce datasets across three categories: (1) medical knowledge, (2) surgical knowledge, and (3) surgical vision-language. 

\subsubsection{Medical Knowledge}

We first aim to build a foundation of medical knowledge. Toward this, we use four instruction fine-tuning datasets: MedMCQA\cite{pal2022medmcqa}, MedQA\cite{jin2021disease}, Medical Flashcards\cite{han2023medalpaca}, and MedInstruct-52k\cite{zhang2023alpacare}. We describe each of these in detail below.

\paragraph{MedMCQA} The MedMCQA\cite{pal2022medmcqa} dataset is a substantial collection designed specifically for medical domain multiple-choice question answering. It comprises of 187,005 high-quality multiple-choice questions in the training set sourced from AIIMS and NEET PG entrance exams, spanning across 2,400 healthcare topics and 21 medical subjects. Each entry in the dataset not only includes a question and its correct answer(s), but also distractor options, requiring reasoning capabilities related to each medical subject. 

\paragraph{MedQA} The MedQA\cite{jin2021disease} dataset comprises open-domain question answering dataset sourced from medical board examinations. It includes questions in English, Simplified Chinese, and Traditional Chinese, with a total of 61,097 questions. For our model we use English questions from this dataset, which were sourced from the US Medical Licensing Exam (USMLE) Step 1 \& Step 2 questions. This dataset is designed to evaluate medical problem-solving abilities on questions that require deep medical knowledge.

\paragraph{Medical Flashcards} The flashcard dataset we use is from the Anki Medical Curriculum which is a subset of the MedAlpaca dataset\cite{han2023medalpaca}. These flashcards were created and maintained by medical students aiming to cover the United States medical school curriculum including subjects such as anatomy, physiology, pathology, and pharmacology. This dataset includes a total of 33,955 sets of QA pairs.

\paragraph{MedInstruct-52k}

The MedInstruct-52K\cite{zhang2023alpacare} dataset consists of 52,000 diverse, synthetically-generated medical instruction-response pairs, developed to improve the instruction-following capabilities of models in the medical domain. The creation of MedInstruct-52K involved collecting an initial set of clinician-designed tasks from various fields such as radiology, genetics, and psychophysiology. These were then used generate a wider array of medical tasks using GPT-4 for in-context learning by randomly selecting three tasks from the initial set and generating twelve tasks during each iteration.

\begin{figure*}
    \centering    \includegraphics[width=1.0\linewidth]{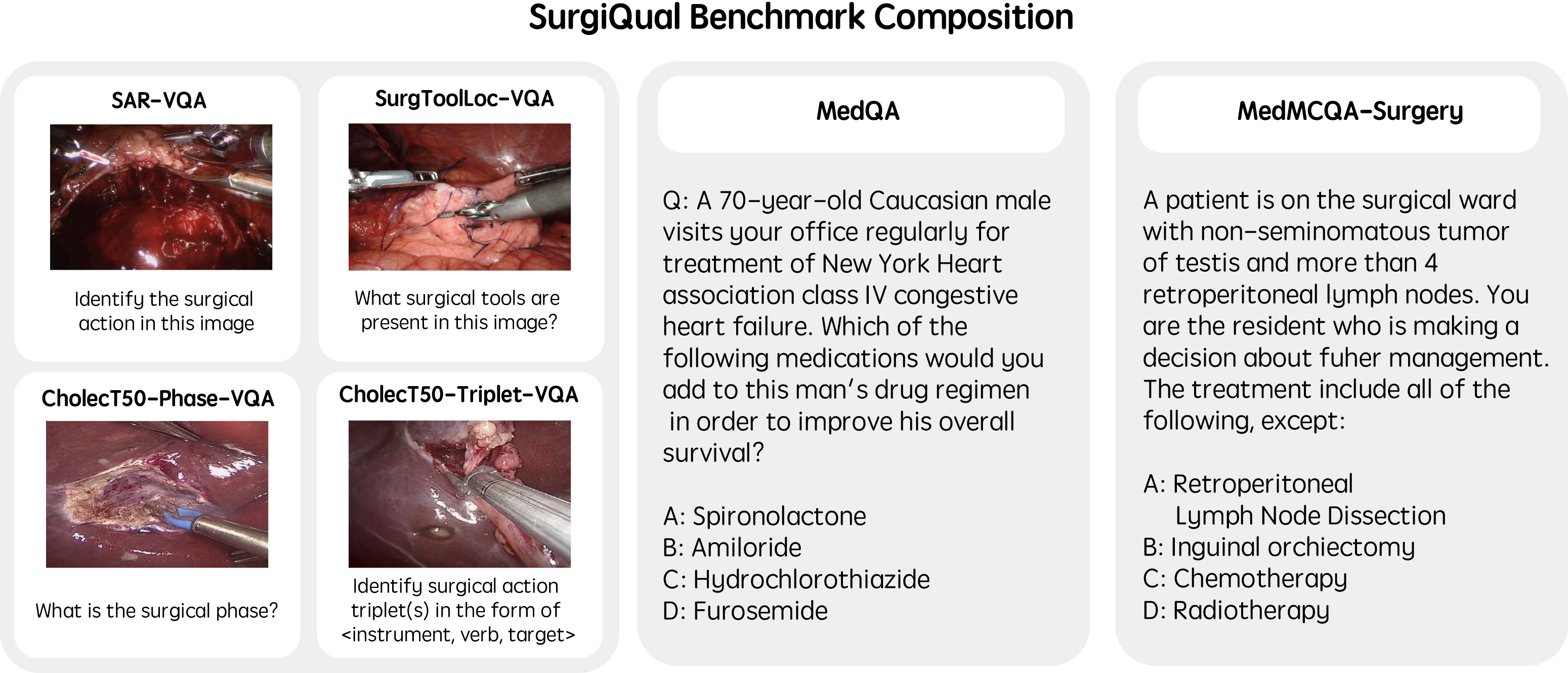}
    \caption{Example questions from each of the six categories of SurgiQual.}
    \label{datasetimg}
\end{figure*}

\subsubsection{Surgical Knowledge}

There is currently a lack of surgery-specific QA datasets designed for training medical language models. To address this deficiency we propose two surgical QA datasets.

\paragraph{SurgTB-QA} SurgTB-QA (SurgTextBook-QA) is a novel dataset comprising textbook question-answer pairs, derived from 445 surgical textbooks and papers. This dataset encompasses a broad spectrum of surgery-specific topics, with an emphasis on general surgery and cardiothoracic surgery. This dataset aims to improve the capabilities of language models in understanding and generating responses related to surgical information. The goal of this dataset is to provide a foundation for understanding concepts in surgery which serve as contextually appropriate responses grounded in factual data sources.

\paragraph{MedMCQA-Surgery} We also curate a set of QA pairs from the MedMCQA dataset\cite{pal2022medmcqa} which contain surgical content. We find that there are 16,862 surgical exam questions from MedMCQA which come in a variety of structures, including the effect of drugs used in surgery, the required next steps in a surgical procedure, or surgical anatomy questions. We train on the training set portion of these questions and use the testing set as part of our SurgiQual benchmark.

\subsubsection{Surgical Vision-Language}

We propose five novel training sets for understanding surgical scenes based on four surgical datasets. These training sets cover a variety of tasks that  are useful for surgery, including recognizing surgical action, phase, triplet tool-action pairs, and tool location. These also include a training set which asks advanced surgical scene questions.

\paragraph{SAR-VQA}

The SAR-VQA dataset is based on the SAR-RARP50 dataset\cite{psychogyios2023sar} and aims to identify surgical actions in a given frame across 11,012 training frames. SAR begins with the question "\textit{Identify the surgical action in this image}" and includes responses relating to the surgical action performed, such as "\textit{The surgical action is pulling the needle out of the tissue}" or "\textit{positioning the needle tip}." We follow the official dataset split, sub-setting 2,882 frames for the test set.

\paragraph{CholecT50-Phase-VQA}
The CholecT50-Phase-VQA is based on the CholecT50\cite{nwoye2022data} dataset which is a series of endoscopic videos of laparoscopic cholecystectomy surgery aimed at describing fine-grained action recognition. This dataset includes a large set of labels, from surgical action triplet to surgical phase recognition.
In the CholecT50-Phase-VQA dataset we use the phase labels and begin the question with "\textit{What is the surgical phase?}"
and responds with the corresponding surgical phase, for instance "\textit{The surgical phase is Calot's triangle dissection.}"
CholecT50-Phase-VQA presents 81,987 questions in the training set and 7,815 questions in the test set. We follow the same dataset split used in the CholecTriplet2021 and CholecTriplet2022 challenges \cite{cholectripletchallenge}.

\paragraph{CholecT50-Triplet-VQA}
The CholecT50-Triplet-VQA dataset is also based on CholecT50\cite{nwoye2022data}. 
Surgical action triplets describe instrument-tissue interactions and are structured as combinations of (instrument, verb, target) and are used in surgery in order to have more precise language for describing surgical scenes. 
In order to request the surgical triplet we ask the following question: "\textit{Identify surgical action triplet(s) in the form of <instrument, verb, target>}" and expects a response such as "\textit{The surgical action triplet(s) are <hook, dissect, gallbladder>,  <grasper, retract, gallbladder>.}" CholecT50-Triplet-VQA presents 11478 questions in the training set.

\paragraph{SurgToolLoc-VQA}
The SurgToolLoc-VQA dataset describes the presence of tools in the surgical scene from a wide variety of different surgeries. Questions from the dataset abide by the following format "\textit{What surgical tools are present in this image?}" and respond with the presented surgical tools, for example "\textit{The surgical tools present are needle driver and cadiere forceps.}" SurgToolLoc-VQA presents 3997 questions in the training set. As the official test set from the SurgToolLoc\cite{zia2023surgical} challenge has not been publicly released, we select 2,472 videos from their train set to serve as our test set.

\paragraph{SynthSSG-VQA}

The SynthSSG-VQA dataset was introduced to improve model generalization and break down the repetitive question-answering structures of previous datasets. Instead of asking the same question with different answers, SynthSSG-VQA asks a wide variety of complex questions not following any particular format, such as "\textit{What is the role of the irrigator in this surgical procedure?}" which has a response of "\textit{The irrigator is likely being used to provide a clear view of the surgical site by removing any blood or debris.}" This dataset was generated by using the SSG-VQA dataset\cite{yuan2024advancing}, which includes a large set of QAs for each surgical image, and inputting the SSG-VQA into GPT-4 to generate meaningful surgical questions about the scene. We opted to not include the original SSG-VQA because the answers were typically 1-2 words (e.g. "\textit{gallbladder}") which qualitatively caused our model to lose its conversational ability.
SynthSSG-VQA presents 1221 questions in the training set.

\subsection{SurgiQual Benchmark}

Current measures of quality for surgical VLMs typically report accuracy on a single classification task (e.g. selecting the phase from five options given an image), where existing language models are taken and adapted with classification heads. 
%In fact, to date there are no existing specialized surgical VLMs for general-purpose problem solving. 
One challenge that the field currently faces as a result of this is that there are also no existing measures of quality for general-purpose surgical VLMs. Toward this, we present a new benchmark, SurgiQual, which evaluates VLMs across a panel of different medical and surgical tasks. Below, we describe the components of SurgiQual.

\paragraph{Medical Exam Performance}
We inherit the MedQA\cite{jin2021disease} question answering test dataset (sourced from USMLE) in order to evaluate basic medical knowledge. We include MedQA because we believe a surgical model having an basic understanding of medicine is an important factor in surgery. We also include questions from the surgery specific test set of the MedMCQA dataset\cite{pal2022medmcqa}. This evaluation includes questions such as which operation would be best in a given context, which tests should be administered for a particular diagnosis, and the definitions of various surgical terms.

\paragraph{Phase Recognition} For evaluating performance for phase recognition, we use the testing set from the CholecT50-Phase-VQA problems. Solutions to this dataset respond to the following question: "\textit{What is the surgical phase?}".

\paragraph{Action Recognition} The surgical action recognition uses 2882 questions from the SAR-VQA testing dataset for evaluation. Solutions to this dataset respond to the following instruction: "\textit{Identify the surgical action in this image}".

\paragraph{Triplet Action Recognition}
The triplet action recognition uses questions from the CholecT50-Triplet-VQA test set for evaluation. This is a more challenging task than action and phase representation because the LLM must identify the instrument, verb, and the target rather than just the action that is being performed. Solutions to this dataset respond to the following instruction: "\textit{Identify surgical action triplet(s) in the form of <instrument, verb, target>}".

\paragraph{Tool Recognition} The tool recognition challenge inherits from the SurgToolLoc-VQA dataset across 2,472 test set images. Solutions to this dataset respond to the following question: "\textit{What surgical tools are present in this image?}".

\subsection{A comparison of models} 

We provide a comprehensive comparison of seven different VLMs in Table \ref{results}. We compare four open-source VLMs from prismatic as well as two closed-source VLMs, GPT-4 and GPT-4 Omni. We compare each model on the six SurgiQual domains, four of which are surgical vision, one on USMLE medical exam questions, and one on surgical exam questions from MedMCQA.

For the text-based benchmarks, MedQA and MedMCQA-Surgery, GPT-4 is the highest performing model on MedQA in the literature. This is also reflected in our results, with GPT-4 obtaining an accuracy of 86.1\% and GPT-4o with 80.5\%. Our model, GP-VLS, obtains the second highest performance on both datasets at 46.1\% (on MedQA) and 52.8\% (on MedMCQA-Surgery). The lowest performing models are the four open-source models in the following order for MedQA and MedMCQA-Surgery respectively: prism-clip+7b (32.3\% and 33.1\%), dinosiglip+7b (34.2\% and 36.0\%), prism-clip+13b (34.6\% and 37.9\%), and dinosiglip+13b (34.3\% and 36.3\%).

For the surgical vision-language benchmarks, GP-VLS clearly outperforms open- and closed-source models for each category. GP-VLS obtains the following percentage improvements over the best model for each category: phase recognition (+8.2\%), triplet recognition (+20.7\%), tool recognition (+14.5\%), and action recognition (+16.9\%).
For phase and tool recognition, GPT-4 Omni and GPT-4 were the highest performing models whereas for triplet and action recognition, dinosiglip+7b and prism-clip+7b were the highest performing respectively.
Overall, there was not a clear pattern between which models were performing the best on each task outside of GP-VLS having the highest accuracy.

We also note that other \textit{surgical} models obtain zero percent accuracy on these benchmarks due to having architectures which only output classification labels instead of text, or having a rigidly defined set of possible text outputs (see Table \ref{datasets}). We were unable to test these models due to the model weights not being open-sourced by the authors, limiting reproducibility.  

%We find that broadly the four open-source language models are the least performant on each of the six metrics, that the closed-source VLMS are the most performant on the text-based exam questions, and 

%\subsection{Clinician quality assessment} 

\section{Discussion}

In this work we introduced a general-purpose vision language model for surgery (GP-VLS) which understands both concepts in medicine and surgery as well as surgical vision-language problems. In order to better measure the quality of surgical VLMs we introduce a comprehensive evaluation across medical and surgical exams (SurgiQual), as well as surgical scene QAs. We also introduce six new training datasets for surgery, with five datasets focusing on vision-language training and one for surgical textbook understanding. Finally, we evaluate seven VLMs on SurgiQual, demonstrating significantly improved performance on surgical vision-language problems and improved performance on medical and surgical knowledge compared with open-source models.

Future work includes further expanding the model's capabilities, such as the ability to produce segmentation maps, train on a wider set of procedures from different medical domains (e.g. urology, cardiothoracic, cranial, etc) and surgical techniques (such as open and laparoscopic). Work in autonomous robotic surgery\cite{kim2024surgical, yu2024orbit, kimlearning, schmidgall2023surgical, moghani2024sufia} could explore training GP-VLS jointly with kinematics recordings in order to develop a vision-language-action (VLA) model\cite{kim2024openvla, ma2024survey}.
Performing experiments in real-world clinical settings will also be important to validate the model's utility in practice and to understand the ways in which surgeons use these tools.

Despite its promising performance, GP-VLS still faces limitations. This may include gaps in knowledge of less common surgical procedures, challenges in interpreting ambiguous visual scenes, and the need for further validation on diverse patient populations. Future surgical VLMs would also benefit from training on surgical videos in addition to image frames in order to capture temporal dynamics, such as the work of SurgVLP\cite{yuan2023learning}. Additionally, practical challenges remain in integrating these AI systems into existing surgical workflows, including the GPU memory requirements and slower inference time.

In conclusion, GP-VLS represents a step forward in the development of general-purpose AI assistants for surgery. By combining medical knowledge with specialized surgical understanding and visual comprehension, it lays groundwork for language-based surgical AI systems. While challenges remain, the potential benefits to surgical practice are considerable. Continued research in this area, guided by rigorous evaluation metrics like SurgiQual and informed by clinical expertise, will be important in realizing the full potential of AI in surgical settings.

\section{Methods}

\subsection{Model Architecture}

We use Llama2-7B-chat\cite{touvron2023llama} as the LLM backbone $f_\phi(\cdot)$ of GP-VLS so that our model has strong chat capabilities. For input image $X_v$, we use the pre-trained CLIP visual encoder ViT-L/14 as $g(\cdot)$ to obtain visual features $Z_v = g(X_v)$. 
A linear layer connects image features to the word embedding space. A trainable projection matrix $W$ converts $Z_v$ into language embedding tokens $H_v$, matching the LLM's word embedding space:

\[
H_v = W \cdot Z_v, \quad \text{with } Z_v = g(X_v)
\]

This produces a sequence of visual tokens $H_v$. We utilize the LLaVA projection scheme\cite{liu2024visual} which enables rapid iteration of data-centric experiments. For each image $X_v$, we generate multi-turn conversation data $(X^1_q, X^1_a, \ldots, X^T_q, X^T_a)$, where $T$ is the number of turns. This is organized as a sequence, where all answers act as the assistant's response. The instruction $X^t_{instruct}$ at the $t$-th turn is defined as:

\[
X^t_{instruct} = 
\begin{cases}
\text{Randomly choose } [X^1_q, X_v] \text{ or } [X_v, X^1_q], & t = 1 \\
X^t_q, & t > 1
\end{cases}
\]

We perform instruction-tuning of the LLM on prediction tokens using its original auto-regressive training objective. For a sequence of length $L$, we compute the probability of target answers $X_a$ by:

\[
p(X_a|X_v, X_{instruct}) = \prod_{i=1}^L p_\theta(x_i|X_v, X_{instruct}, x_{<i})
\]

where $\theta$ are the trainable parameters. We use $X_{instruct,*}$ for readability.

\bibliography{sample}
\end{document}